\documentclass{article}

\usepackage{PRIMEarxiv}

\usepackage[utf8]{inputenc} 
\usepackage[T1]{fontenc}    
\usepackage{hyperref}       
\usepackage{url}            
\usepackage{booktabs}       
\usepackage{amsfonts}       
\usepackage{nicefrac}       
\usepackage{microtype}      
\usepackage{lipsum}
\usepackage{fancyhdr}       
\usepackage{graphicx}       
\graphicspath{{media/}} 
\usepackage{xcolor}
\usepackage{algorithm}
\usepackage{amsmath}

\DeclareMathOperator*{\argmin}{arg\,min}
\usepackage{caption} 
\usepackage{hyperref}

\usepackage{subfig}

\title{Negotiated Representations to Prevent Overfitting in Machine Learning Applications 
}

\author{
  Nuri Korhan \\
  Istanbul Technical University  \\
  Maslak, Istanbul, Turkey \\
  \texttt{\ korhan@itu.edu.tr} \\
   \And
  Samet Bayram \\
  University of Delaware \\
  Newark, De, USA\\
  \texttt{sbayram@udel.edu} \\
}

\begin{document}
\maketitle

\begin{abstract}
Overfitting is a phenomenon that occurs when a machine learning model is trained for too long and focused too much on the exact fitness of the training samples to the provided training labels and cannot keep track of the predictive rules that would be useful on the test data. This phenomenon is commonly attributed to memorization of particular samples, memorization of the noise, and forced fitness into a data set of limited samples by using a high number of neurons. While it is true that the model encodes various peculiarities as the training process continues, we argue that most of the over-fitting occurs in the process of reconciling sharply defined membership ratios. In this study, we present an approach that increases the classification accuracy of machine learning models by allowing the model to negotiate output representations of the samples with previously determined class labels. By setting up a negotiation between the model’s interpretation of the inputs and the provided labels, we not only increased average classification accuracy but also decreased the rate of over-fitting without applying any other regularization tricks. By implementing our negotiation paradigm approach to several low regime machine learning problems by generating over-fitting scenarios from publicly available data sets such as CIFAR 10, CIFAR 100, and MNIST we have demonstrated that the proposed paradigm has more capacity than its intended purpose. We are sharing the experimental results and inviting the machine-learning community to explore the limits of the proposed paradigm. We also aim to incentive the community to exploit the negotiation paradigm to overcome the learning-related challenges in other research fields such as continual learning. The Python code of the experimental setup is uploaded to Git Hub. \footnote{GitHub: \url{https://github.com/nurikorhan/Negotiated-Representations}.}
\end{abstract}


\section{Introduction}
The general structure of the supervised deep learning \cite{lecun2015deep} requires us to rely on the labels provided by humans beforehand. These provided labels create a cost function that informs the model on how far off the prediction is. By trying to decrease the cost with the help of back-propagation, the model is expected to encode the underlying input-output relationships into its weights. This approach works extraordinarily well for big data regimes. However, it becomes unreliable in low data regimes. If the model is large enough to ensure the fitness between samples and their respective labels, it tends to encode the aspects of the samples that are irrelevant to the classification process. The burden of encoding the irrelevant features renders the classifier less accurate in interpreting the test samples. This phenomenon is called over-fitting \cite{li2019research}. Researchers have developed highly sophisticated methods to prevent the data from over-fitting. 

Data augmentation and regularization techniques are limited in compensating for the over-fitting problem \cite{balestriero2022effects}. Data augmentation aims to increase the number of samples by slightly vibrating the training samples in the multi-dimensional space to allow each sample to represent its corresponding neighborhood, and regularization limits the movement of the weights by adding weight punishment to the loss function. None explicitly addresses the problem of not having membership ratios of the samples to their classes. While it is true that a sample belongs to a class or not, it is impossible to justifiably represent the samples by their corresponding labels if the proximity of the individual samples to all classes is not accounted for. This is also the case for humans. If the object is sufficiently distant from an observer, the observer will assign equal probabilities for each class. In other words, "If it is far enough, it can be anything." As the object gets closer, the observer gradually changes the class probabilities (say, it looks like a dog, but it could still be anything). Furthermore, if human errors are also considered, it should be kept in mind that the output representations should never be exact. To our knowledge, no study in the literature has proposed an algorithm that computationally scrutinizes the provided labels in supervised learning. In this respect, ANFIS \cite{jang1993anfis} was an early attempt to break the ice on the quantification of class memberships. ANFIS was proposed for increasing the speed of learning in back-propagation-based algorithms.  However, a pre-encoded knowledge base (rule base and database) requires a deep understanding of the supervision criteria for determining the membership functions. Needless to mention the cost. 

In this study, we attempt to deal with over-fitting by setting up a negotiation between the model’s interpretation of the input samples and the provided labels. So that the model’s belief can be gradually injected into the output labels. The amount of the model’s belief injected into the labels at any iteration is determined by a variable called negotiation rate. By gradually increasing the negotiation rate, we can ensure that as the model obtains a better fitness to the labels, it is rewarded with a better position in the negotiation table. Therefore, the model comes to better fitness and does not spend much energy accounting for wrong labels, exceptions (outliers), and aforementioned membership values that are justified by the quality of the observations. Also, gradually scrutinizing the categorical labels relieves us of endless hours of fitting the data into a paradigm.

The motivation for this study is rooted in the exploration of generic and specific differences in representations \cite{williams2013gilles}, as well as Ludwig Wittgenstein's philosophical investigations \cite{wittgenstein2010philosophical}. When attempting to identify an object, a child must examine the object through various dimensions, such as visual properties (e.g., shape, color, texture, size) and the object's function (e.g., taste, content, or purpose). Nonetheless, not all dimensions are consistently accessible for object recognition, and even when all the required information is available, it is not always feasible to employ every dimension for differentiating between classes. This phenomenon necessitates a careful examination of class memberships. 

This study aims to tackle the over-fitting problem in low data regime machine learning problems by setting up a negotiation between the model's belief of training labels and the labels provided by human supervisors. Our proposed method aims to balance the model's assumptions and human input, creating a more robust and accurate classifier, even when dealing with limited data

The organization of the paper is as follows: {\color{black}{In section II we describe the experimental setups that are used in simulations.}} In section III, simulation results are presented. The section IV includes discussions and possible future directions for the proposed algorithm.

\section{The Method}
\label{sec:proposed}

\subsection{Over-fitting and Prevention Strategies }
Over-fitting \cite{dietterich1995overfitting} is a phenomenon that occurs when the model is trained for too long and focused too much on the exact fitness of the training samples to the provided training labels and cannot keep track of the predictive rules that would be useful on the test data. In literature, over-fitting is commonly attributed to memorization of the particular samples, noise, and other peculiarities of data samples by using high number of neurons. While it is true that the model also encodes undesired aspects of the data samples as training process continues, we argue that most of the over-fitting occurs in the process of reconciling sharply defined membership ratios to specific classes.

The loss of the individual differences in hierarchical systems \cite{deleuze1977capitalism} is also of great significance in understanding representation learning. Although neural networks can be considered hierarchical systems, individual differences may not be lost but transformed into the means of compensation for the inconvenient membership ratios. However, the main concern of the aforementioned problem is related to the fidelity of the representation to the actual sample, and individual differences should be filtered out for higher representational capacity in the face of a particular objective such as classification. 

 The third argument that should be discussed is that the certainty of a decision depends highly on the quality of the observation. For instance, let’s assume that we try to recognize a cat or a dog by using a picture given. If the distance of the animal from the camera is long enough to cover all distinctive differences between cats and dogs, we decide that the probabilities are the same. As the camera gets closer to the animal of interest, the distinctive differences start to appear, and the probability of the sample belonging to one of the classes increases. Trying to form an association among poorly represented memberships may cause the network to encode the exceptions and, therefore inject specific and undesired noise into the principal components of the distinction process. 
 Too prevent machine learning models from over-fitting, several methods have been proposed: dropout \cite{srivastava2014dropout}, L1 \cite{park2007l1}, L2 \cite{cortes2012l2}, and augmentation \cite{balestriero2022effects}. To our knowledge, none of the proposed methods in the literature computationally scrutinizes the provided labels in supervised learning.

In this article, we suggest that a sharp definition of the membership ratios may be the leading source of over-fitting.To prevent over-fitting, we propose enabling the model to negotiate the membership ratios of all samples to all classes by slightly adjusting the provided labels, such as changing a label from 1 to 0.98, to better represent the sample's relationship with the rest of the data set. To test our hypothesis, we have generated a number of over-fitting scenarios and allowed the model to compensate for the lack of precision in the provided labels. We have tested the proposed training paradigm on publicly available benchmark datasets such as CIFAR10, CIFAR100, MNIST, and Fashion MNIST. In order to generate a low data regime, we have selected a small set of training and test examples for each dataset. The results on all datasets have shown that the negotiation between the model and the provided labels is a powerful method in preventing over-fitting. 

\subsection{Negotiated Representations}

The general structure of supervised learning requires us to rely on the labels provided by humans beforehand. These provided labels are then used to create a cost function that informs the model on how far off the prediction is. By trying to decrease the cost with the help of back-propagation, the model is expected to encode the underlying input-output relationships into its weights. This logic works extraordinarily well for big data regimes. However, it becomes unreliable in low data regimes as the size of the samples increases. If the model is large enough to ensure the fitness between samples and their respective labels, it tends to encode the aspects of the samples that are irrelevant to the classification process. We can represent the neural network as a mapping function as: 

\begin{equation}\label{eq:supervised_learning}
    \boldsymbol{Y}: f(\boldsymbol{X},\boldsymbol{\theta},\boldsymbol{b}),
\end{equation}
where $\boldsymbol{X}$
represents the data instances in the training set, $\boldsymbol{Y}$ represents ground truth labels, $\boldsymbol{b}$ is bias, and $\theta$ represents the network parameters.

The network parameters are updated at each epoch with back-propagation depending on the predicted labels, $Y'=f(X)$ and the loss function $L: J(\boldsymbol{Y},\boldsymbol{Y'})$, where $J$ represents the cost function. The optimization of network parameters is shown as:

\begin{equation}\label{eq:standard_tr}
    \boldsymbol{\theta^*} = \argmin_{\theta \in \Theta}\frac{1}{L}\sum_{i=1}^L J(y_i,y'_i).
\end{equation}

\begin{figure}[h]
  \centering
\includegraphics[width=.7\textwidth]{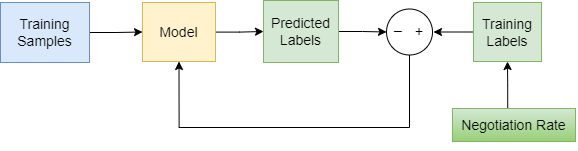}
  \caption{The model diagram with negotiation rate.}
  \label{fig1}
\end{figure}

In this study, we attempt to deal with over-fitting by setting up a negotiation between the model’s interpretation of the inputs and the provided labels. So that the model’s belief will be gradually injected into the data set itself. The amount of the model’s belief that is injected into the labels is determined by a variable called negotiation rate denoted by 'n'. By gradually increasing the negotiation rate, we ensure that as the model obtains a better fitness to the labels, it is rewarded with a better position at the negotiation table. Therefore, the model reaches a better fitness and does not spend much energy for the sake of encoding the exceptions and individual identities of the samples. A closed form of the proposed model is shown in Fig \ref{fig1}.

where negotiated labels are calculated by a weighted average of predicted labels and original labels. When we include the negotiation rate in the training process, our training loss function becomes as:
\begin{equation}\label{eq:supervised}
    \boldsymbol{L}: J(\boldsymbol{(1-n)\cdot Y},\boldsymbol{n \cdot Y'}).
\end{equation}

\begin{figure}
  \centering
\includegraphics[width=.7\textwidth]{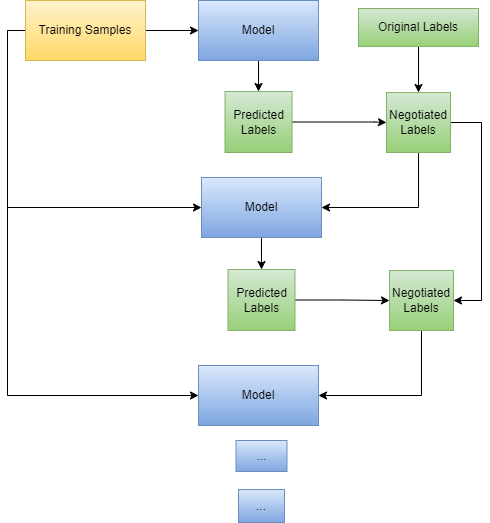}
  \caption{Flowchart of the model.}
  \label{fig2}
\end{figure}

Thus, the optimization term shapes as:

\begin{equation}\label{eq:standard_tr}
    \boldsymbol{\theta^*_{nr}} = \argmin_{\theta \in \Theta}\frac{1}{L}\sum_{i=1}^L J(\boldsymbol{(1-n)\cdot y_i},\boldsymbol{n \cdot y_i'}).
\end{equation}
It should be kept in mind that, at the end of each negotiation phase original labels are switched with negotiated labels that are calculated in that phase. Furthermore, since the model gains more confidence as training continues, the negotiation rate is also linearly increased at the end of each epoch. This change means that the coefficient of the model's predictions will increase and the coefficient of the provided labels will decrease at the end of each negotiation phase. The linear increment in the negotiation rate limits the number of negotiations that take place throughout training. Otherwise, negotiations would arrive at a point where the model's coefficient in the weighted average (n) would be more than 1, and the previously determined labels' coefficient (1-n) would go below zero. A detailed flowchart of the model is given in Figure \ref{fig2}.

\section{Experiments}
\label{sec:experiments}
\subsection{The Network Structure}
\label{subsec:network_structure}

In order to evaluate the performance of the proposed paradigm, we designed four different models. We provided the models with sufficient data to draw meaningful conclusions while limiting the amount of data used for training in order to induce over-fitting. The experimental setups described below serve as a proof of concept and demonstrate the behaviors of the models. One downside of exceptionally high success rates in classifiers is that they can be attributed to the injection of test data set information into the model through hyper-parameter tuning. Optimizing each part of the model for maximum test performance may also result in encoding many peculiarities of the test data set within the model. Consequently, building upon any paradigm requires us to reverse the optimization process for a more objective evaluation of the method. For this reason, we found it more beneficial to focus solely on demonstrating the behavior of the model throughout the experiments.

In the context of investigating overfitting induction and its mitigation through the implementation of various algorithms, we employed distinct configurations of convolutional neural networks for each dataset. 
For all of the convolution layers within the networks, we utilized the Rectified Linear Unit (ReLU) activation function, as it offers several advantages, such as reduced likelihood of the vanishing gradient problem and improved training speed. In contrast, for the final fully connected layer in each model, we employed the soft-max layer, as it enables the output to be constrained between the range of 0 and 1, which is particularly useful for the deployment of the negotiation paradigm in classification tasks.

\subsection{Simulation Results and Discussion}

This section presents evidence of the effectiveness of the proposed method for preventing over-fitting in the model. First, a comprehensive analysis of the results is provided, including figures and their interpretations. Second, a comparison is made between the baseline model and the model trained with the proposed paradigm. Specifically, Table \ref{table:Table_loss} summarizes the performance metrics of the two models. Figure \ref{fig:mnist_nr} shows that the phenomenon observed is similar to Figure \ref{fig:fminsit_nr}, which confirms the efficacy of the proposed method. Additionally, Figure \ref{fig:cifar10_nr} and Figure \ref{fig:cifar100_nr} present the results of the model trained on the Cifar-10 and Cifar-100 datasets respectively. The findings are promising, even though some aspects remain unexplained. In Table \ref{table:Table_acc}, we provide testing accuracy performances of the baseline model and proposed model.

\begin{table}[H]
\caption{Accuracy comparison of baseline model and the proposed model on test data.}

\centering

\begin{tabular}{ccccc}
Dataset \ & Baseline Model\ & Proposed Model &  \\ \cline{1-4}
MNIST          & 0.828     & 0.867                                    &  \\ 
Fashion MNIST  & 0.719      & 0.766                                  &  \\ 
CIFAR10        & 0.326    & 0.343                                    &  \\ 
CIFAR100       & 0.460       & 0.491                                    &  \\  \cline{1-4}
\end{tabular}
\label{table:Table_acc}
\end{table}
The comparison between the losses of the baseline model and the proposed model is demonstrated in Table \ref{table:Table_loss}.

\begin{table}[H]
 \caption{Loss comparison of baseline model and proposed model}
 \centering
\begin{tabular}{ccccc}
Dataset \ & Baseline Model\ & Proposed Model &  \\ \cline{1-4}
MNIST          & 0.92     & 0.41                                    &  \\ 
Fashion MNIST  & 1.94      & 0.78                                 &  \\ 
CIFAR10        & 4.48  & 2.13                                    &  \\ 
CIFAR100       & 13.43       & 5.18                                   &  \\  \cline{1-4}
\end{tabular}
\label{table:Table_loss}
\end{table}

The proposed method outperforms the baseline model for each data set. We provide plots of training and validation accuracies with an increasing number of epochs in later sections to visualize over-fitting and model training performances.

\subsubsection{MNIST}

\begin{figure}[h]
    \centering
    \subfloat[\centering Loss]{{\includegraphics[width=7cm]{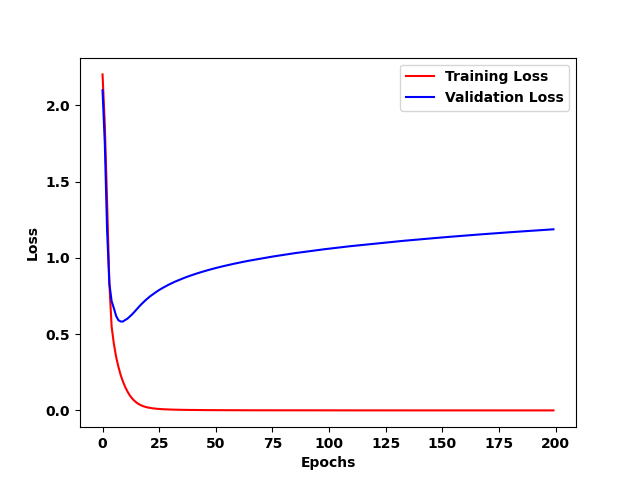} }}
    \qquad
    \subfloat[\centering Accuracy]{{\includegraphics[width=7cm]{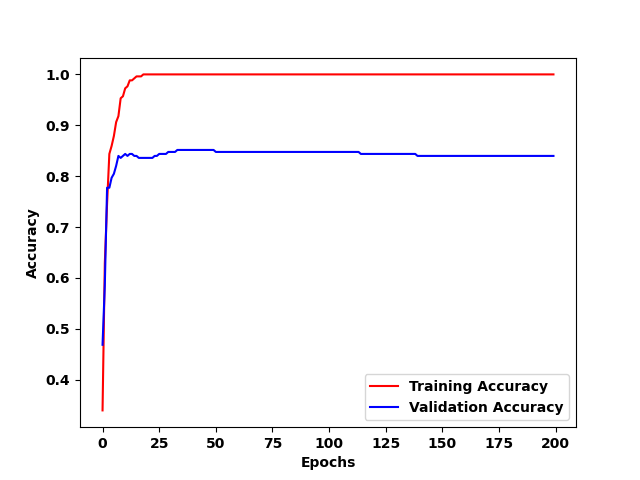} }}
    \caption{Standard Network Performance on MNIST data set}%
    \label{fig:mnist_normal}%
\end{figure}

We constructed a model for MNIST data set that consists of two convolutional layers, each having 32 and 64 filters respectively, followed by a fully-connected layer containing ten neurons. The training set consisted of 256 samples, and the test set contained 256 samples. Due to the low number of samples and the simplicity of recognizing digits, it was relatively easy to generate an over-fitting scenario. Moreover, there were no high-level relationships that could prevent the model from accurately classifying the samples. Additionally, since the images are single-channeled gray images, there were no color complications. The accuracy and loss values for the model are provided in Figure \ref{fig:mnist_normal}.

\begin{figure}[h]
    \centering
    \subfloat[\centering Loss]{{\includegraphics[width=7cm]{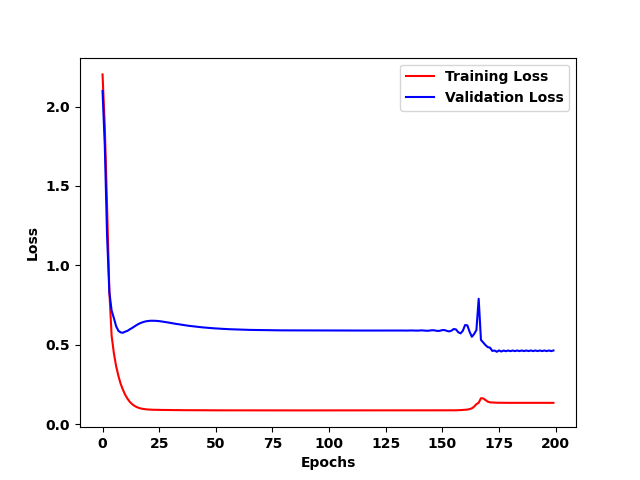} }}
    \qquad
    \subfloat[\centering Accuracy]{{\includegraphics[width=7cm]{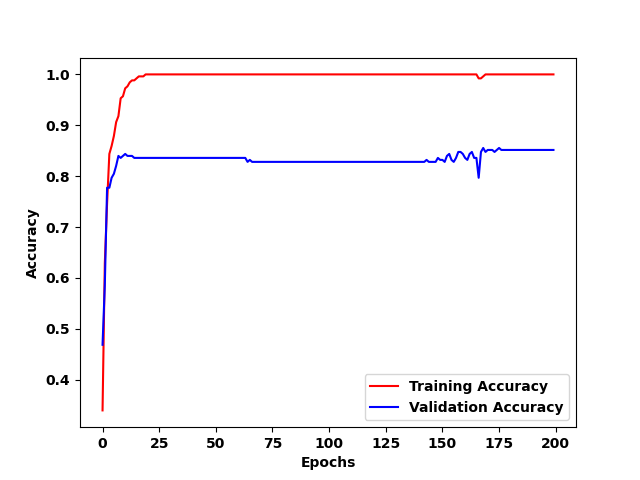} }}
    \caption{Performance of Network with Negotiated Representation on MNIST data set}%
    \label{fig:mnist_nr}%
\end{figure}

As observed in Figure \ref{fig:mnist_normal}-a, the model starts over-fitting after around ten epochs. When we applied the proposed method to the model we observed that the over-fitting was reduced and the accuracy was improved as it is seen in Figure \ref{fig:mnist_nr}.

\subsubsection{Fashion-MNIST}

\begin{figure}[h]
    \centering
    \subfloat[\centering Loss]{{\includegraphics[width=7cm]{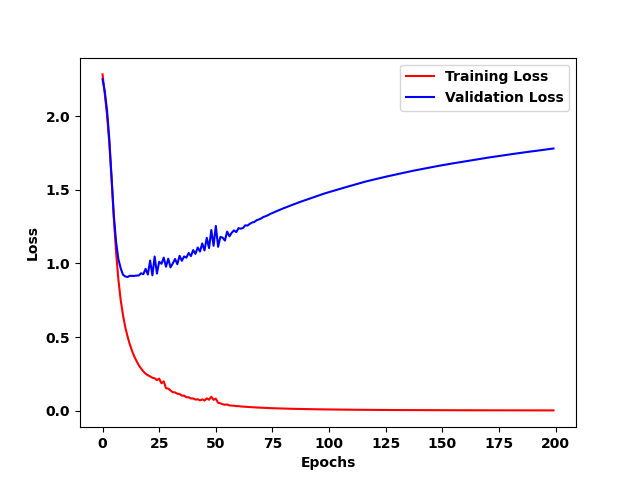} }}
    \qquad
    \subfloat[\centering Accuracy]{{\includegraphics[width=7cm]{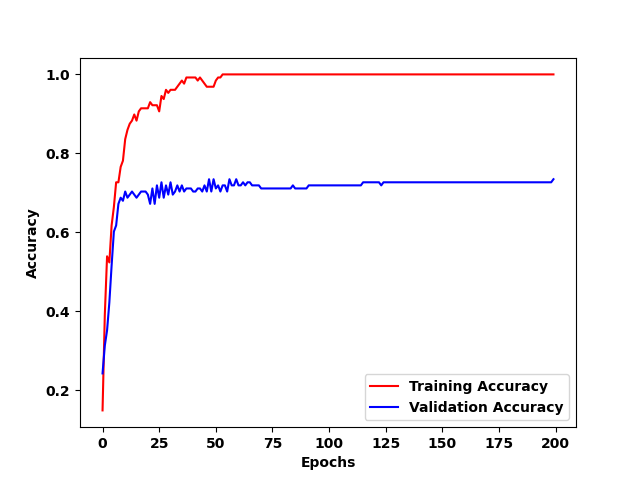} }}
    \caption{Standard Network Performance on Fashion MNIST data set}%
    \label{fig:fmnist_normal}%
\end{figure}

The model for Fashion MNIST data set has the same network parameters as we used for training MNIST  data set. The training set consisted of 128 samples, and the test set comprised 128 samples.  The accuracy and loss values for the model are provided in Figure \ref{fig:fmnist_normal}.

\begin{figure}[h]
    \centering
    \subfloat[\centering Loss]{{\includegraphics[width=7cm]{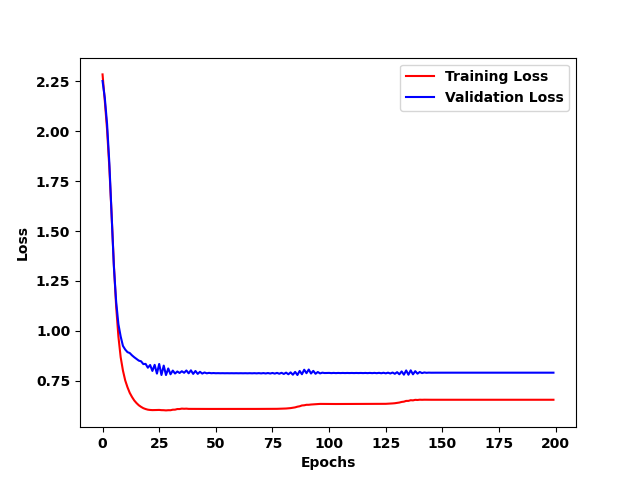} }}
    \qquad
    \subfloat[\centering Accuracy]{{\includegraphics[width=7cm]{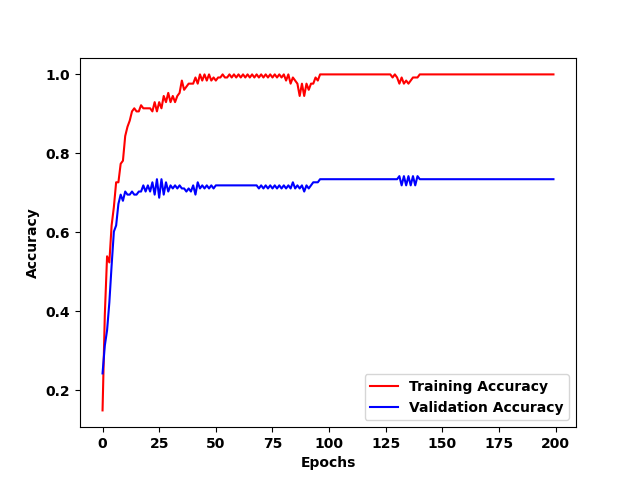} }}
    \caption{Performance of Network with Negotiated Representation on Fashion-MNIST data set}%
    \label{fig:fminsit_nr}%
\end{figure}

As we noticed from Figure \ref{fig:fmnist_normal}-a, the model is heavily over-fitted. The model improves after the proposed negotiation representation regarding loss and accuracy as it is seen in Figure \ref{fig:fminsit_nr}.

\subsubsection{CIFAR-10}
\begin{figure}[h]
    \centering
    \subfloat[\centering Loss]{{\includegraphics[width=7cm]{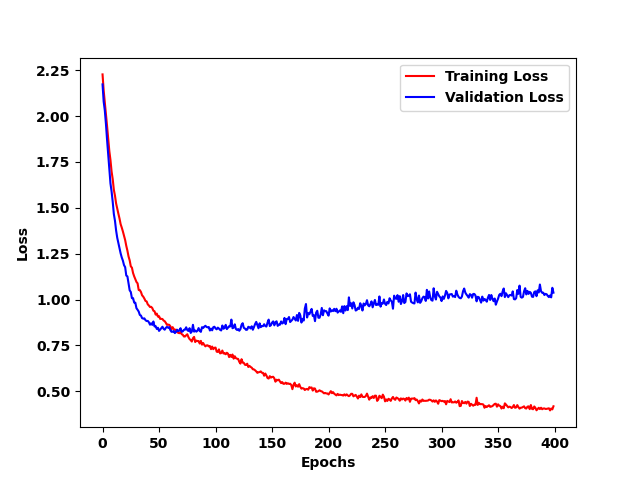} }}
    \qquad
    \subfloat[\centering Accuracy]{{\includegraphics[width=7cm]{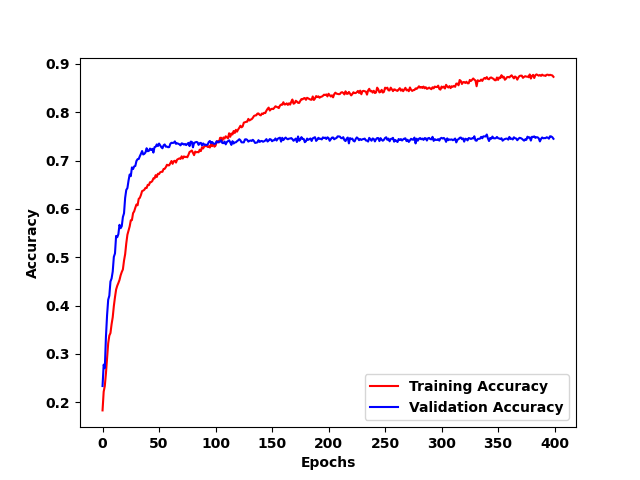} }}
    \caption{Standard Network Performance on CIFAR-10 dataset}%
    \label{fig:cifar10_normal}%
\end{figure}

Creating an over-fitting scenario for the CIFAR-10 data set proved to be more challenging than for MNIST and Fashion MNIST. This increased difficulty can be attributed to the higher-level relationships and color images present in the data set, resulting in three channels of information per sample, which adds complexity to the learning task, and generating an over-fitting scenario with a small network becomes more difficult. Figure \ref{fig:cifar10_normal} shows the loss and accuracy performance of regular training on Cifar10 data set. The observed results clearly demonstrate that the proposed paradigm significantly reduced the validation data set's loss. However, the increase in test accuracy was relatively minor and not indicative of the improvement in test loss. Nevertheless, any improvement is beneficial in the context of machine learning.

\begin{figure}[h]
    \centering
    \subfloat[\centering Loss]{{\includegraphics[width=7cm]{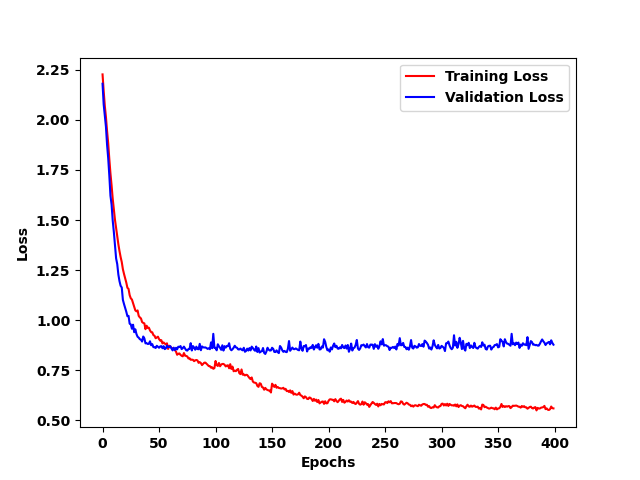} }}
    \qquad
    \subfloat[\centering Accuracy]{{\includegraphics[width=7cm]{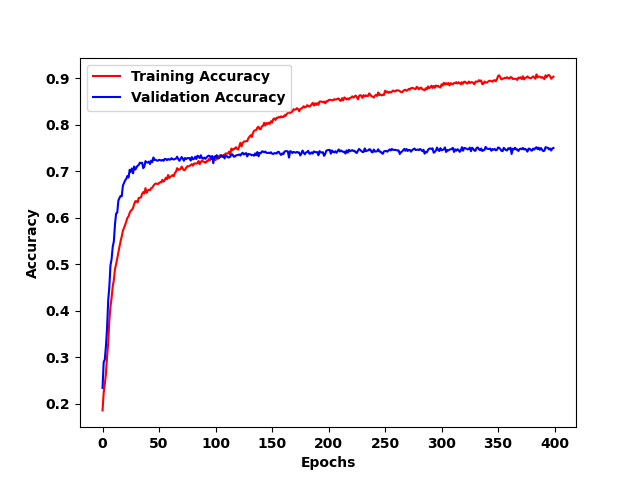} }}
    \caption{Performance of Network with Negotiated Representation on CIFAR-10 data set}%
    \label{fig:cifar10_nr}%
\end{figure}

Similar to the previous simulations, we observe over-fitting by evaluating the loss and accuracy plots. After training with the negotiation paradigm, we obtained an improvement in the loss and accuracy of the model as demonstrated in Figure \ref{fig:cifar10_nr}.

\subsubsection{CIFAR-100}
For the CIFAR-100 data set, we designed a more complex model due to the large number of classes. The model comprises six convolutional layers, each containing 64, 64, 128, 128, 256, and 256 filters, respectively, followed by a fully-connected layer with 512 neurons and a soft-max layer. The training set included 45,000 samples, while the test set consisted of 5,000 samples. A more extensive model is necessary for managing the increased number of classes, and fitting such a model requires a larger data set. However, a larger data set can make achieving fitness more challenging. Our choice of model and data set size was based on these considerations, as our objective was to first generate over-fitting and then mitigate it using our proposed paradigm.

\begin{figure}[h]
    \centering
    \subfloat[\centering Loss]{{\includegraphics[width=7cm]{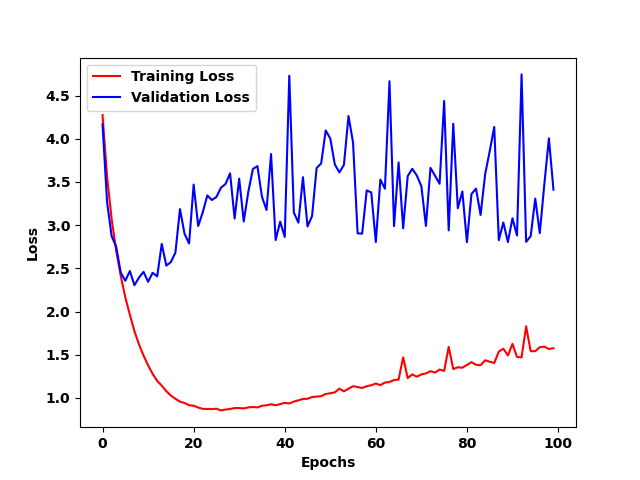} }}
    \qquad
    \subfloat[\centering Accuracy]{{\includegraphics[width=7cm]{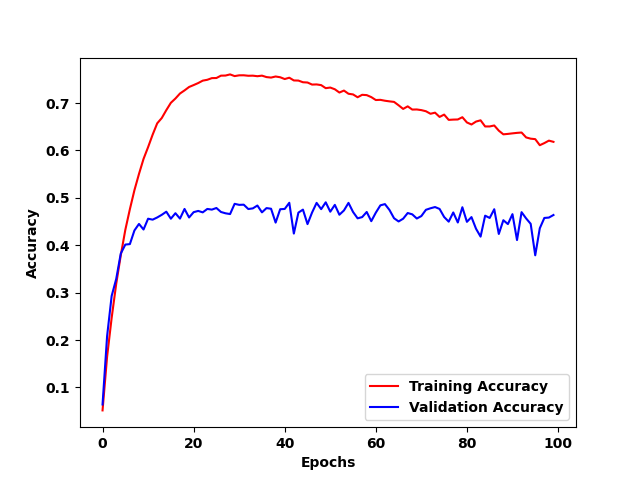} }}
    \caption{Standard Network Performance on CIFAR-100 data set}%
    \label{fig:cifar100_normal}%
\end{figure}

Generating an over-fitting scenario for the CIFAR-100 data set proved more difficult than for other data sets due to the large number of classes, complex relationships within the data, and the use of colored images containing three channels of information. In the simulations for the CIFAR-100 data set, we found that decreasing the loss was relatively manageable while improving the accuracy was more challenging. Consequently, representational fidelity does not always guarantee high accuracy. In this scenario, over-fitting is unavoidable. To mitigate over-fitting, we incorporated dropout and max-pooling layers. Figure \ref{fig:cifar100_normal} shows the loss and accuracy performance of the model. 

\begin{figure}[h]
    \centering
    \subfloat[\centering Loss]{{\includegraphics[width=7cm]{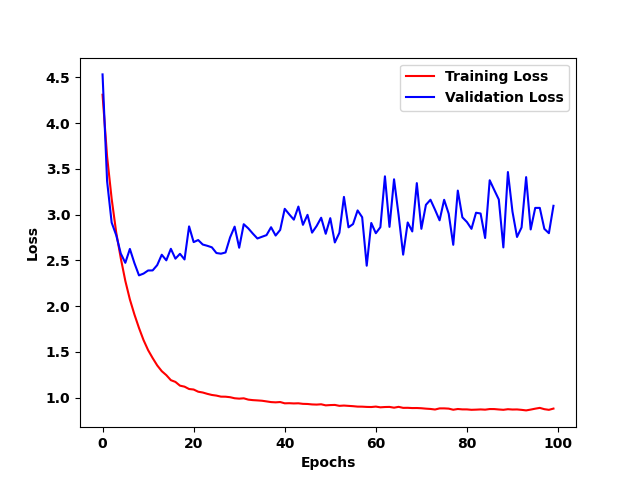} }}
    \qquad
    \subfloat[\centering Accuracy]{{\includegraphics[width=7cm]{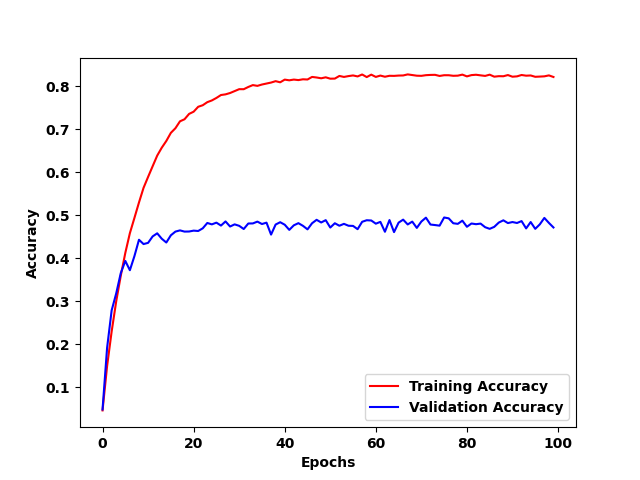} }}
    \caption{Performance of Network with Negotiated Representation on CIFAR-100 data set}%
    \label{fig:cifar100_nr}%
\end{figure}

Figure \ref{fig:cifar100_normal} shows an obvious over-fitting after a few epochs. Over-fitting is improved significantly after using negotiation representation along the model as it is seen in Figure \ref{fig:cifar100_nr}. 

\section{Conclusion and Future Work}
In this study, we have presented a novel algorithm to mitigate over-fitting in classification tasks, particularly in low-data regimes. The method has been applied to several data sets, including MNIST, Fashion MNIST, CIFAR 10, and CIFAR 100, demonstrating its potential to address a broad range of low-data regime challenges. The success of the method, however, is dependent on the negotiation rate, and further research is required to investigate the relationship between the data set and the optimal negotiation rate for the best performance. We aim to draw the attention of the machine learning community towards developing novel methods for justifying assigned labels. We propose that a significant discrepancy between training and test loss could stem from the fact that the provided labels are not adequately justified by the characteristics of the observations. The justification will likely be context-dependent. Considering the context of the data set, each deviation from the most optimal representations should be injected into labels as class memberships. Doing so will enhance the model's performance and provide a more philosophically sound justification for deep learning applications.

We also believe that the negotiated learning paradigm holds great promise for continual learning, offering a more efficient, intuitive, and sustainable approach compared to current methods in the literature \cite{salami2021state}. By injecting the model's past experiences into future labels, one can potentially mitigate catastrophic forgetting to a new degree without compromising the plasticity of the neurons or relying on memory-intensive replay scenarios. It can also be coupled with existing paradigms to update the state-of-the-art performances. In order to not break the flow of this study, we did not share any particular experimental setup. Our method might have already achieved state-of-the-art performance in class incremental continual learning, utilizing a variant of the negotiation algorithm. Although we will not disclose any specific details until the completion of our ongoing research, we feel obligated to note that negotiated representations may have a far greater reach than preventing over-fitting in machine learning applications.

\bibliographystyle{unsrt}  
\bibliography{references}  
\end{document}